\documentclass{article}

\usepackage[utf8]{inputenc}
\usepackage[T1]{fontenc}
\usepackage{hyperref}
\usepackage{url}
\usepackage{booktabs}
\usepackage{amsfonts}
\usepackage{amsmath}
\usepackage{amssymb}
\usepackage{nicefrac}
\usepackage{xcolor}
\usepackage{graphicx}
\usepackage{subcaption}
\usepackage{multirow}
\usepackage{algorithm}
\usepackage{algorithmic}
\usepackage{enumitem}
\usepackage{tikz}
\usetikzlibrary{arrows.meta, positioning}
\usepackage{pgfplots}
\pgfplotsset{compat=1.18}
\usepackage{natbib}
\usepackage[margin=1in]{geometry}

\newcommand{\corecraft}{\textsc{Corecraft}}

\title{EnterpriseBench \corecraft: Training Generalizable Agents on High-Fidelity RL Environments
}

\author{   Sushant Mehta\thanks{Correspondence to: \texttt{sushantmehta@surgehq.ai}}, Logan Ritchie, Suhaas Garre, Ian Niebres, Nick Heiner, Edwin Chen  \\   
Surge AI \\   
}

\date{}  

\begin{document}

\maketitle

\begin{abstract}
We show that training AI agents on high-fidelity reinforcement learning environments produces capabilities that generalize beyond the training distribution. We introduce \corecraft{}, the first environment in EnterpriseBench, Surge AI's suite of agentic RL environments. \corecraft{} is a fully operational enterprise simulation of a customer support organization, comprising over 2,500 entities across 14 entity types with 23 unique tools, designed to measure whether AI agents can perform the multi-step, domain-specific work that real jobs demand. Frontier models such as Claude Opus 4.6, GPT-5.2, and Gemini 3.1 Pro solve fewer than 35\% of tasks when all expert-authored rubric criteria must be satisfied. Using this environment, we train GLM~4.6 with Group Relative Policy Optimization (GRPO) and adaptive clipping. After a single epoch of training, the model improves from 25.37\% to 36.76\% task pass rate on held-out evaluation tasks. More importantly, these gains transfer to out-of-distribution benchmarks: +4.5\% on BFCL Parallel, +7.4\% on $\tau^2$-Bench Retail, and +6.8\% on Tool Decathlon (Pass@1). We believe three environment properties are consistent with the observed transfer: task-centric world building that optimizes for diverse, challenging tasks; expert-authored rubrics enabling reliable reward computation; and enterprise workflows that reflect realistic professional patterns. Our results suggest that environment quality, diversity, and realism are key factors enabling generalizable agent capabilities.
\end{abstract}

\section{Introduction}
\label{sec:intro}

The deployment of AI agents in production settings remains fairly limited despite rapid capability improvements on research benchmarks. A recent survey of 306 practitioners found that 68\% of deployed agents execute ten or fewer steps before human intervention, with reliability cited as the primary development challenge~\citep{pan2025measuring}. This gap between benchmark performance and deployment readiness suggests that current training approaches may not fully develop the robust, transferable skills required for real-world operation.

We hypothesize that this gap stems partly from the characteristics of training environments. Many existing agent benchmarks use simplified simulations, synthetic data, or contrived task structures that fail to capture the complexity of real-world workflows~\citep{zhou2024webarena,xie2024osworld}. When agents are trained on such environments, they may learn environment-specific heuristics rather than generalizable problem-solving strategies.

This paper presents evidence supporting an alternative approach: training on high-fidelity environments designed around realistic enterprise workflows. We introduce results from \corecraft{}, a reinforcement learning environment simulating a customer support agent at a PC components company. \corecraft{} was developed for training and evaluating frontier models, where it revealed systematic capability gaps and an empirically-derived hierarchy of agentic capabilities~\citep{ritchie2026hierarchy}. In this work, we demonstrate that realistic environments also serve as effective training substrates. \corecraft{} optimizes for task quality through three core design principles:

\begin{enumerate}[leftmargin=*,nosep]
    \item \textbf{Task-centric design}: Every entity, tool, and data source exists to support diverse, challenging tasks rather than to maximize world complexity.
    \item \textbf{Expert-authored evaluation}: Domain experts design both tasks and detailed rubrics, enabling reliable automated reward computation.
    \item \textbf{Realistic workflows}: Tasks mirror genuine professional patterns including multi-step reasoning, constraint handling, and structured communication.
\end{enumerate} 

\vspace{5pt}
Using this environment, we train GLM~4.6 with GRPO and demonstrate that a single epoch of training produces substantial improvements that transfer to external benchmarks. Our contributions are:
\vspace{5pt}
\begin{itemize}[leftmargin=*,nosep]
    \item We demonstrate +11.39 percentage point improvement on held-out \corecraft{} tasks, exceeding the capability gap between Claude Sonnet 4.5 and Claude Opus 4.5 (+7.05 pp).
    \item We show transfer to out-of-distribution benchmarks: +4.5\% on BFCL Parallel function calling, +7.4\% on $\tau^2$-Bench Retail customer service, and +6.8\% on Tool Decathlon long-horizon tool use.
    \item We provide qualitative analysis of learned behaviors, identifying three categories of improvement: multi-step workflow execution, constraint handling, and response quality.
    \item We present evidence that environment quality, diversity, and realism are important factors enabling generalization, and discuss implications for agent training methodology.
\end{itemize}

\section{Related Work}
\label{sec:related}

\subsection{Agent Benchmarks and Evaluation}

The landscape of agent evaluation has evolved from simplified web interfaces~\citep{liu2018miniwob,shi2017world} toward realistic, execution-based environments. WebArena~\citep{zhou2024webarena} introduced self-hostable web environments with 812 long-horizon tasks. OSWorld~\citep{xie2024osworld} extended evaluation to desktop environments across operating systems. AppWorld~\citep{trivedi2024appworld} provides a controllable environment for benchmarking interactive coding agents across 9 apps and 457 APIs.

For software engineering, SWE-bench~\citep{jimenez2024swebench} evaluates agents on 2,294 real GitHub issues, while Terminal-Bench~\citep{merrill2026terminal} focuses on command-line tasks spanning scientific workflows, network configuration, and data analysis. TheAgentCompany~\citep{xu2024theagentcompany} simulates a software company with 175 diverse professional tasks requiring web browsing, coding, and communication with simulated coworkers, finding that best agents complete only 24\% of tasks autonomously. This consistent finding across benchmarks underscores the challenge of building reliable agents for real-world deployment.

Customer service evaluation has been addressed by $\tau$-bench~\citep{yao2024tau} and its successor $\tau^2$-bench~\citep{barres2025tau2}, which introduced the pass$^k$ metric measuring reliability across multiple attempts. Tool Decathlon or Toolathlon~\citep{li2025toolathlon} benchmarks agents on 108 diverse, long-horizon tasks spanning 32 software applications and 604 tools, with execution-based evaluation. AgentBoard~\citep{ma2024agentboard} provides multi-turn evaluation with subgoal progress tracking across diverse agentic tasks.

Recent surveys have organized this growing landscape. \citet{yehudai2025survey} provide a comprehensive taxonomy across four dimensions: fundamental capabilities, application-specific benchmarks, generalist agents, and evaluation frameworks. Our work contributes to this literature by demonstrating that high-fidelity environments designed for evaluation can also serve as effective training substrates.

\subsection{Tool Use and Function Calling}

Tool use represents a critical capability for agents interacting with external systems. The Berkeley Function Calling Leaderboard (BFCL)~\citep{patil2025bfcl} established systematic evaluation across 2,200+ test cases covering serial and parallel function calls. Gorilla~\citep{patil2024gorilla} demonstrated that specialized training on API documentation can surpass GPT-4 on function calling tasks.

ToolLLM~\citep{qin2024toolllm} introduced ToolBench with 16,464 real APIs and proposed depth-first search reasoning for complex tool sequences. T-Eval~\citep{chen2024teval} decomposed tool utilization into six sub-processes for fine-grained capability diagnosis. These works highlight that tool use involves not just API knowledge but also planning, error recovery, and output formatting.

\subsection{Reinforcement Learning for Language Model Agents}

Modern RL approaches for language models trace from proximal policy optimization (PPO)~\citep{schulman2017ppo} through InstructGPT's three-stage training~\citep{ouyang2022instructgpt} to more recent methods. Direct Preference Optimization (DPO)~\citep{rafailov2023dpo} simplified training by eliminating explicit reward models, though subsequent work found PPO often outperforms DPO when properly tuned~\citep{ivison2024unpacking}.

Group Relative Policy Optimization (GRPO)~\citep{shao2024deepseekmath} eliminates the critic network by estimating baselines from group scores, significantly reducing memory requirements. DAPO~\citep{yu2025dapo} introduced adaptive clipping and other techniques preventing entropy collapse during training. DeepSeek-R1~\citep{deepseek2025r1} demonstrated that reasoning capabilities can emerge through pure RL with verifiable rewards.

Recent work has begun extending RL training to agentic settings. AgentTuning~\citep{zeng2024agenttuning} demonstrated that hybrid instruction-tuning on agent trajectories can enhance agent capabilities without degrading general abilities. FireAct~\citep{chen2024fireact} showed that fine-tuning language models on GPT-4 agent trajectories yields substantial performance gains, though primarily through supervised learning rather than RL. The ReAct framework~\citep{yao2023react} demonstrated the value of interleaving reasoning and acting, enabling agents to generate verbal reasoning traces alongside actions. Reflexion~\citep{shinn2023reflexion} showed that agents can improve through verbal self-reflection, using linguistic feedback from failed attempts to inform subsequent trials. Agent-R1~\citep{cheng2025agentr1} proposed a modular RL framework for training LLM agents with end-to-end reinforcement learning on multi-step tasks. ToolRL~\citep{qian2025toolrl} provided the first comprehensive study of reward design for tool-use RL, demonstrating that GRPO-based training achieves 17\% improvement over base models and critically showing that RL-trained models generalize to unfamiliar tool-use scenarios where supervised fine-tuning struggles. In instruction following, rubric-based reward signals have shown promise: \citet{he2025rifl} introduced RIFL, a pipeline that uses expert-curated rubrics as reward signals for RL training, achieving substantial improvements on instruction-following benchmarks. Our work applies a similar rubric-based reward approach to agentic tasks in realistic enterprise environments, demonstrating that this paradigm extends beyond instruction following to multi-step tool-use workflows.

\section{The \corecraft{} Environment}
\label{sec:environment}

\subsection{Design Philosophy}

\corecraft{} simulates a customer support agent at Corecraft Computers, Inc., a fictional online PC parts retailer. The environment provides a stateful world in which agents interact with databases, tools, and simulated customers to complete support tasks.

Our central design principle is that high-quality, diverse tasks \textbf{provide a useful training signal}; everything else exists to support task diversity and difficulty. This contrasts with approaches that solely maximize entity or tool counts without sufficient complexity or diversity.

We prefer worlds with realistic entities that support more diverse, challenging tasks to worlds with many entities supporting redundant or trivial tasks.

\subsection{World Structure}

The environment is packaged as a self-contained Docker bundle, enabling stateful interactions where the container maintains world state (orders, tickets, inventory) for the duration of each episode. The bundle includes:

\begin{itemize}[leftmargin=*,nosep]
    \item \textbf{Entities}: Customer records, order histories, product catalogs, support tickets, shipping information, and company policies stored in structured JSON format.
    \item \textbf{Tools}: Database queries, order management APIs, communication interfaces, and diagnostic utilities exposed via Model Context Protocol (MCP)~\citep{anthropic2024mcp}, enabling standardized tool invocation across different agent frameworks.
    \item \textbf{Tasks}: Customer support scenarios with associated rubrics, each containing prompts, system context, and verifiable evaluation criteria.
\end{itemize}

Tasks range from simple lookups to complex multi-step workflows requiring coordination across multiple data sources. The MCP server handles all tool calls, routing agent requests to the appropriate backend operations while maintaining transactional consistency.

\subsection{Task Categories}

Tasks span several categories of different difficulty levels:

\paragraph{Information Retrieval:} Basic tasks requiring agents to locate and present information from databases, such as order status queries or product specifications.

\paragraph{Communication:} Tasks requiring professional customer-facing responses that synthesize the retrieved information into clear, actionable messages.

\paragraph{Reasoning:} Tasks requiring the application of business rules, temporal constraints, or policy conditions. For example, determining refund eligibility based on purchase date, return window, and product condition.

\paragraph{Multi-Step Workflows:} Complex tasks requiring sequential operations in the correct order: validate input, identify issues, apply fixes, compute totals, and format output.

\subsection{Rubric Design}

Each task includes expert-authored rubrics that enable automated evaluation. Rubrics decompose success into verifiable criteria, such as:

\begin{itemize}[leftmargin=*,nosep]
    \item \textbf{Completeness}: Did the agent address all required aspects?
    \item \textbf{Correctness}: Are factual claims accurate given the world state?
    \item \textbf{Constraint Satisfaction}: Are business rules and policies correctly applied?
    \item \textbf{Format Compliance}: Does the output follow required structure?
\end{itemize}
\vspace{5pt}
This decomposition enables reward computation for RL training while providing diagnostic information about failure modes. Task-level pass rates require \emph{all} rubric criteria to be satisfied, providing a strict measure of end-to-end task completion.

\section{Frontier Model Performance on \corecraft{}}
\label{sec:frontier}

To characterize the difficulty of \corecraft{} and establish the headroom available for training, we evaluated frontier models on the current \corecraft{} benchmark task set. Table~\ref{tab:frontier} reports task pass rates, where a task is counted as passed only when all rubric criteria are satisfied across trajectories. These results reveal substantial headroom and even the strongest model achieves only 30.80\% task pass rate on this set.

\begin{table}[t]
\centering
\begin{tabular}{lc}
\toprule
\textbf{Model} & \textbf{Task Pass Rate (\%)} \\
\midrule
Claude Opus 4.6 (Adaptive Thinking + Max Reasoning Effort) & 30.80 \\
GPT-5.2 (High Reasoning Effort) & 29.70 \\
Gemini 3.1 Pro & 27.20 \\
Claude Opus 4.6 (High Reasoning Effort) & 26.20 \\
DeepSeek v3.2 Thinking & 24.10 \\
Claude Opus 4.6 & 22.10 \\
Grok 4.1 Fast Reasoning & 20.50 \\
GPT-5.2-Codex (xHigh Reasoning Effort) & 20.10 \\
Gemini 3 Flash & 20.00 \\
GLM-5 & 17.40 \\
Claude Sonnet 4.6 & 16.40 \\
Gemini 3 Pro & 14.40 \\
Qwen 3.5 Plus & 11.3 \\
Nova 2 Pro (High Reasoning Effort) & 8.90 \\
Kimi K2.5 & 8.70 \\
Mistral Large 3 & 3.60 \\
Qwen3 Max Thinking & 3.60 \\
\bottomrule
\end{tabular}
\vspace{5pt}
\caption{Frontier model performance on the current \corecraft{} benchmark task set. Pass rate requires all rubric criteria to be satisfied. Even the best model (Claude Opus 4.6 Adaptive) solves fewer than 31\% of tasks.}
\label{tab:frontier}
\end{table}

\subsection{Illustrative Frontier Model Failures}

Analysis of frontier model trajectories reveals recurring failure patterns that persist even in the strongest models. We highlight three representative examples.

\paragraph{Poor Search Strategy.} Effective tool use requires not just invoking the correct tool, but formulating queries that retrieve the most relevant information. In one task, a customer asks about an issue with a monitor they purchased. GPT-5.2 immediately begins searching the knowledge base for general troubleshooting articles about monitors, but fails to find the correct article through this broad approach. The more effective strategy, which few model trajectories achieved, was to first retrieve the customer's actual order and product details, then search the knowledge base for articles specific to that product. This failure reflects a common pattern: models default to generic keyword searches rather than reasoning about which tool calls would most efficiently narrow the search space. A human agent would naturally look up the specific product first before searching for troubleshooting guidance.

\paragraph{Failure to Paginate Through Incomplete Results.} A recurring failure across multiple frontier models involves accepting incomplete search results without attempting pagination. In one task, GPT-5.2 is asked to find all orders for a product in November. The search tool returns a maximum of 10 results per call, and GPT-5.2 is clearly aware of this limit, as it explicitly sets the limit parameter. However, when the tool returns exactly 10 results (going back only to November 18th), GPT-5.2 accepts these as complete and reports its answer. The model fails to recognize the strong contextual signal that exactly hitting the result limit suggests truncation, and does not issue follow-up queries to retrieve remaining results. This pattern recurs across models: while tools do not explicitly indicate that results are incomplete, there is sufficient contextual information to infer incompleteness. The failure suggests that current models struggle to react to contextual cues they were not necessarily expecting, even when those cues are straightforward.

\paragraph{Incomplete Exploration of Available Tools.} When asked to determine the price of a PC build, Claude Opus 4.6 uses only the \texttt{searchBuilds} tool to estimate the cost by summing individual component prices. It does not consider using \texttt{searchProducts} to check whether the same configuration is available as a prebuilt product at a lower price. Other model trajectories successfully discover this alternative path. This failure reflects a broader pattern: models often anchor on the first plausible tool for a task and do not explore whether alternative tools might yield better results. In a realistic support setting, a human agent would routinely check both custom and prebuilt options before quoting a price.

\vspace{5pt}
These failures illustrate that \corecraft{} tasks expose genuine capability gaps in frontier models across three dimensions: search strategy (formulating effective queries rather than defaulting to generic searches), persistence and contextual understanding (recognizing implicit signals of incomplete results and paginating accordingly), and comprehensive tool exploration (considering all relevant tools rather than anchoring on the first viable option). These are precisely the types of skills that benefit from RL training with detailed rubric feedback, as the rubrics explicitly penalize inefficient information retrieval, incomplete results, and missed alternatives.

\section{Training Methodology}
\label{sec:method}

We train a frontier-scale language model using reinforcement learning on \corecraft{}, using the environment's expert-authored rubrics as verifiable reward signals. The goal is twofold. First, we aim to improve performance on the held-out \corecraft{} evaluation set, demonstrating that rubric-based RL training produces measurable gains on challenging agentic tasks. Second, and more importantly, we evaluate whether these gains \textit{transfer} to external benchmarks that the model was never trained on, testing our hypothesis that high-fidelity environments teach generalizable agentic skills rather than environment-specific heuristics.

\subsection{Base Model and Data}

We use GLM~4.6 as our base model, a 357B parameter mixture-of-experts architecture with 32B active parameters. We used a random train-test split from the \corecraft{} environment with 1000 tasks used for training and 150 tasks in the held-out evaluation test.

\subsection{Training Infrastructure}

\begin{figure}[t]
\centering
\begin{tikzpicture}[
    box/.style={rectangle, draw, minimum width=2.2cm, minimum height=1cm, align=center, font=\small},
    container/.style={rectangle, draw, dashed, inner sep=12pt},
    arrow/.style={->, >=Stealth},
    biarrow/.style={<->, >=Stealth}
]

\node[container, minimum width=10cm, minimum height=2.4cm] (miles) at (0, 0) {};

\node[box] (training) at (-3, 0) {Training\\(Megatron)};
\node[box] (buffer) at (0, 0) {Data Buffer\\(Bridge)};
\node[box] (rollout) at (3, 0) {Rollout\\(SGLang)};

\draw[biarrow] (training) -- (buffer);
\draw[biarrow] (buffer) -- (rollout);

\node[container, minimum width=5.2cm, minimum height=2.2cm] (docker) at (2.5, -4.2) {};
\node[font=\small, anchor=north] at (2.5, -3.1) {Corecraft Docker Container};

\node[box, minimum width=1.6cm] (tools) at (1.2, -4.7) {Tools/\\Schemas};
\node[box, minimum width=1.6cm] (entities) at (3.8, -4.7) {Entities/\\World DB};

\node[box, minimum width=4.8cm] (judge) at (2.5, -7.5) {LLM Judge (Verifier)\\Rubric-based grading};

\draw[arrow] (rollout.south) -- (3, -3);

\draw[arrow] (2.5, -5.4) -- node[right, font=\small] {Trajectory} (judge.north);

\draw (judge.west) -- (-1.5, -7.5) -- (-1.5, 0) -- (buffer.west);
\draw[arrow] (-1.5, -1) -- (-1.5, 0) -- (buffer.west);
\node[font=\small, anchor=south] at (-0.5, -7.5) {Rewards};

\node[font=\small, align=left, anchor=east] at (-1.7, -3.5) {Trajectories\\+ Rewards};

\draw (training.west) -- (-4.8, 0) -- (-4.8, -1.8) -- (3, -1.8);
\draw[arrow] (3, -1.8) -- (rollout.south);
\node[font=\small, anchor=east] at (-5, 0) {Weight Sync};

\end{tikzpicture}
\vspace{6pt}
\caption{Architecture of the RL training loop with \corecraft{}. The rollout engine generates agent responses, routing tool calls to stateful Docker containers running the MCP server. Completed trajectories are evaluated by an LLM judge against task rubrics, with rewards flowing back to the training loop.}
\label{fig:architecture}
\end{figure}
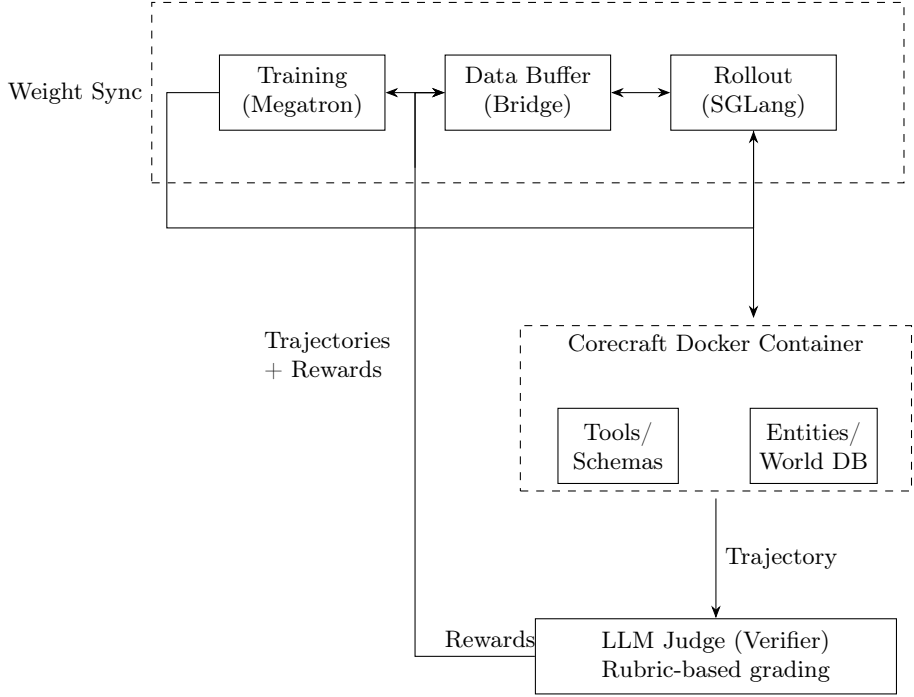

Figure~\ref{fig:architecture} illustrates the overall architecture. The training pipeline consists of three stages operating in a continuous loop:

\paragraph{Rollout Generation:} For each training prompt, we generate 16 rollouts. Each rollout interacts with its own stateful \corecraft{} Docker container. The rollout engine generates model responses, with tool calls routed to the Model Context Protocol (MCP) server running inside the container. The container maintains world state (orders, tickets, inventory) for the duration of the rollout, enabling multi-turn trajectories that proceed until the agent produces a final response.

\paragraph{Reward Computation:} Once a trajectory completes, the final response is evaluated against task-specific rubrics using an LLM judge. Each rubric criterion is a verifiable assertion (e.g., ``The response should identify that the PSU wattage is insufficient for the GPU's requirements''). Rewards are computed as the proportion of satisfied criteria.

\paragraph{Training Update:} Rollout results (trajectories and rewards) are written to a data buffer. The Megatron training loop reads batches from the buffer, computes policy gradients using GRPO, and updates model weights. Updated weights are synchronized back to the SGLang rollout engine for the next iteration.

\subsection{GRPO with Adaptive Clipping}

We train using Group Relative Policy Optimization~\citep{shao2024deepseekmath} with modifications inspired by DAPO~\citep{yu2025dapo}. GRPO eliminates the critic network by computing advantages relative to other completions in a group. 

\subsection{Reward Computation}

Rewards are computed from rubric evaluations using an LLM judge. Each task completion receives scores on individual criteria, aggregated into a continuous score for training:

\begin{equation}
r = \frac{1}{|C|} \sum_{c \in C} \mathbf{1}[\text{criterion } c \text{ satisfied}]
\end{equation}

where $C$ is the set of rubric criteria. This provides a dense reward signal while maintaining interpretable task-level metrics.

\paragraph{Judge Reliability.} We use an LLM as the grading mechanism, the rubric criteria are designed to be granular and objective, substantially reducing the ambiguity that can undermine LLM judge reliability. Each task is decomposed into several concrete, factual criteria that test a specific, verifiable assertion grounded in the environment state. 





\subsection{Environment Architecture}

A key distinction between \corecraft{} and existing agentic benchmarks is the scale and interconnectedness of its world state. Where benchmarks such as $\tau^2$-Bench~\citep{barres2025tau2} provide curated conversation scenarios and Toolathlon~\citep{li2025toolathlon} offers diverse but largely independent task environments, \corecraft{} simulates a single, coherent enterprise with the full complexity of a real operational backend. The environment contains over 2,500 entities spanning 14 entity types (customers, orders, products, builds, support tickets, SLAs, shipping records, product compatibility rules, warranty policies, loyalty tiers, knowledgebase articles, promotions, inventory levels, and company policies), all with realistic interdependencies. A single customer query may require the agent to traverse connections across orders, products, compatibility data, warranty terms, and active promotions before arriving at a correct response.

Agents interact with this world through 23 unique tools exposed via MCP, ranging from database queries (\texttt{searchOrders}, \texttt{searchProducts}) to state-modifying operations (\texttt{updateTicketStatus}, \texttt{processReturn}) and diagnostic utilities (\texttt{validateBuildCompatibility}). Crucially, the environment includes the ``noise'' of real enterprise data: conflicting timestamps, incomplete records, pagination limits that do not signal truncation, and edge cases that require policy-aware reasoning. Models that perform well on static benchmarks often fail here in consequential ways, such as issuing incorrect refunds, entering infinite tool-call loops, or exposing customer PII through poorly constructed queries.

The training bundle is packaged as a self-contained Docker image and is designed to be framework-agnostic, providing all components needed for RL integration, as seen in Table~\ref{tab:bundle}

\begin{table}[h]
\centering
\small
\begin{tabular}{lll}
\toprule
\textbf{Component} & \textbf{Location} & \textbf{Purpose} \\
\midrule
Tasks & \texttt{tasks/*.json} & Prompts and rubrics for training/eval \\
MCP Server & \texttt{server/} & Handles tool calls from the agent \\
World Data & \texttt{entities/} & 2,500+ entities across 14 types \\
System Prompts & \texttt{system-prompts/} & Agent context and policies \\
Tool Definitions & \texttt{mcp-tools.json} & 23 tools with schemas \\
\bottomrule
\end{tabular}
\caption{Components of the \corecraft{} Docker bundle for RL integration.}
\label{tab:bundle}
\end{table}

This bundle can integrate with any RL framework supporting MCP tool-use, multi-turn trajectory collection, and custom reward functions. The stateful Docker container maintains transactional consistency across multi-turn episodes, ensuring that tool calls that modify world state (e.g., updating a ticket or processing a return) are reflected in subsequent queries within the same trajectory.

\section{Results}
\label{sec:results}

\subsection{In-Distribution Performance}

Table~\ref{tab:main_results} presents task pass rates on the held-out \corecraft{} evaluation set used during training.\footnote{Since training was conducted, new tasks have been added to \corecraft{} benchmark, resulting in the Corecraft benchmark set reported in Section~\ref{sec:frontier}. The results in Table~\ref{tab:main_results} reflect the task split available at training time.} A task is counted as passed only when \emph{all} rubric criteria are satisfied, making this a strict measure of end-to-end task completion. After one epoch of training, GLM~4.6 improves from 25.37\% to 36.76\%, a gain of 11.39 percentage points.

\begin{table}[t]
\centering
\begin{tabular}{lc}
\toprule
\textbf{Model} & \textbf{Pass Rate (\%)} \\
\midrule
GPT-5.1 High & 36.86 \\
\textbf{GLM 4.6 (1 Epoch)} & \textbf{36.76} \\
Claude Opus 4.5 & 33.49 \\
GPT-5 & 30.45 \\
Claude Sonnet 4.5 & 26.44 \\
\textbf{GLM 4.6 (Baseline)} & \textbf{25.37} \\
Gemini 3 Pro Preview & 24.84 \\
GPT-5.2 & 24.77 \\
Claude Haiku 4.5 & 7.21 \\
\bottomrule
\end{tabular}
\vspace{5pt}
\caption{Task pass rates on the held-out evaluation set used during training. Pass rate requires all rubric criteria to be satisfied. The trained model surpasses Claude Opus 4.5 and approaches GPT-5.1 High performance.}
\label{tab:main_results}
\end{table}

\subsection{Qualitative Analysis of Learned Behaviors}

To understand what the model learned, we analyzed paired trajectories comparing baseline and trained model behavior on identical tasks. Three categories of improvement emerged:

\paragraph{Multi-Step Workflow Execution.} The most significant improvement was in end-to-end execution with correct task decomposition. The trained model follows intended operation sequences: validate constraints, identify incompatibilities, apply fixes, compute totals, present structured output.

\textit{Example}: A PC build compatibility task requires verifying component compatibility, identifying issues, recommending fixes, and computing updated totals. The baseline performs partial compatibility checks with incorrectly sequenced recommendations. The trained model executes the correct workflow: (1) retrieve order and catalog data, (2) identify case-motherboard incompatibility first, (3) recommend case replacement in correct sequence, (4) compute accurate price totals.

\paragraph{Constraint Handling and Reasoning.} The trained model significantly improved at applying constraints correctly: filtering data by time windows, handling ``before/after'' relationships, and correctly joining records based on ordering.

\textit{Example}: A task requires finding orders placed after a support ticket creation within a 30-day window with specific status constraints. The baseline retrieves orders but applies inconsistent time comparisons. The trained model uses explicit datetime comparisons, filters to post-ticket orders only, enforces the 30-day window, and produces clean structured output with timestamps.

\paragraph{Response Quality and Structure.} The trained model produces more complete, actionable outputs aligned with realistic customer communication patterns.

\textit{Example}: A shipment status email task requires generating professional communication with carrier details and next steps. The baseline produces inconsistent phrasing without clear guidance. The trained model produces structured email with: (1) order reference, (2) current shipment status with carrier name, (3) expected delivery window, (4) clear customer next steps, (5) support contact information.

\vspace{5pt}

These three competencies are not specific to PC components or Corecraft's particular tools, and they help explain the observed transfer to external benchmarks, as discussed in the next section. Because \corecraft{} tasks mirror realistic enterprise workflows, the model learns general professional patterns rather than environment-specific shortcuts.

\subsection{Out-of-Distribution Generalization}

Table~\ref{tab:ood_results} presents results on benchmarks outside the \corecraft{} training distribution. The trained model shows consistent improvements across long-horizon tool-use benchmarks.

\begin{table}[t]
\centering
\begin{tabular}{lccc}
\toprule
\textbf{Benchmark} & \textbf{Baseline} & \textbf{Trained} & \textbf{$\Delta$} \\
\midrule
BFCL Parallel & 91.0\% & 95.5\% & +4.5\% \\
BFCL Simple & 91.5\% & 93.5\% & +2.0\% \\
$\tau^2$-Bench Retail & 68.7\% & 76.1\% & +7.4\% \\
Toolathlon & 18.8\% & 25.6\% & +6.8\% \\
\bottomrule
\end{tabular}
\vspace{6pt}
\caption{Out-of-distribution benchmark results showing transfer from \corecraft{} training. Improvements are consistent across function calling, customer service, and long-horizon tool-use benchmarks.}
\label{tab:ood_results}
\end{table}

\paragraph{BFCL Results.} The Berkeley Function Calling Leaderboard~\citep{patil2025bfcl} evaluates parallel and simple function calling scenarios. The trained model improves by 4.5\% on parallel calls, which require invoking multiple APIs correctly in a single turn, suggesting better tool coordination learned from \corecraft{}'s multi-tool workflows.

\paragraph{$\tau^2$-Bench Results.} The $\tau^2$-Bench Retail domain~\citep{barres2025tau2} evaluates customer service conversations requiring database lookups, policy application, and response generation. The 7.4\% improvement demonstrates that customer support skills transfer across different retail contexts.

\paragraph{Toolathlon Results.} Toolathlon~\citep{li2025toolathlon} evaluates language agents on 108 diverse, long-horizon tasks spanning 32 software applications and 604 tools, requiring an average of approximately 20 interaction turns per task. The trained model improves from 18.8\% to 25.6\% (+6.8\%), placing it between Claude Haiku 4.5 and Grok-4 in our evaluations. This is a particularly notable transfer result because Toolathlon tasks (e.g., managing Kubernetes deployments, grading assignments on Canvas, syncing product inventories across databases) are substantially different from \corecraft{}'s customer support domain, suggesting that the trained model acquired broadly applicable agentic skills rather than domain-specific shortcuts.

\subsection{Detailed Toolathlon Analysis}
\label{sec:toolathlon_detail}

Given that Toolathlon represents the more challenging benchmark in our evaluation suite, we provide a detailed analysis of performance to characterize where and how \corecraft{} training improves long-horizon agentic capabilities.

\paragraph{Evaluation Protocol.} We evaluated the fine-tuned model on all 108 Toolathlon tasks across 3 independent runs using the Toolathlon private evaluation service. We report three complementary metrics following~\citet{li2025toolathlon} and the pass$^k$ formulation introduced by~\citet{barres2025tau2}: \textbf{Pass@1} (mean pass rate across runs, measuring expected single-attempt performance), \textbf{Pass@3} (fraction of tasks passed on at least one run, measuring best-case coverage), and \textbf{Pass$^3$} (fraction of tasks passed on all runs, measuring worst-case reliability).

\paragraph{Overall Results.} Table~\ref{tab:toolathlon_overall} summarizes the aggregate improvements. Pass@1 increases by 6.8 percentage points (from 18.8\% to 25.6\%), with per-run pass rates of 25.0\%, 26.9\%, and 25.0\%, indicating high stability across runs. The reduction in standard deviation from $\pm$2.2\% (baseline) to $\pm$0.6\% (trained) reflects more consistent performance. Pass$^3$ nearly doubles from 9.3\% to 17.6\%, indicating that the trained model not only solves more tasks overall but does so more reliably: 19 tasks (17.6\%) are solved consistently across all 3 runs, compared to 10 (9.3\%) for the baseline.

\begin{table}[t]
\centering
\begin{tabular}{lccc}
\toprule
\textbf{Metric} & \textbf{Base} & \textbf{Trained} & \textbf{$\Delta$} \\
\midrule
Pass@1 & 18.8 $\pm$ 2.2\% & 25.6 $\pm$ 0.6\% & +6.8 pp \\
Pass@3 & 29.6\% & 35.2\% & +5.6 pp \\
Pass$^3$ & 9.3\% & 17.6\% & +8.3 pp \\
\bottomrule
\end{tabular}
\vspace{5pt}
\caption{Toolathlon aggregate results across 3 independent runs.}
\label{tab:toolathlon_overall}
\end{table}

\paragraph{Performance by Task Category.} Table~\ref{tab:toolathlon_categories} breaks down Pass@1 by task category. The trained model shows improvements across most categories, with the largest absolute gains in categories that require structured data manipulation and multi-step web interactions.

\begin{table}[t]
\centering
\small
\begin{tabular}{lrccc}
\toprule
\textbf{Category} & \textbf{Tasks} & \textbf{Pass@1} & \textbf{Pass@3} & \textbf{Pass$^3$} \\
\midrule
Data Analysis / Finance & 15 & 37.8\% & 53.3\% & 20.0\% \\
Web / Forms / Documents & 10 & 36.7\% & 50.0\% & 30.0\% \\
E-Commerce / WooCommerce & 11 & 30.3\% & 36.4\% & 18.2\% \\
Canvas / Education & 15 & 28.9\% & 33.3\% & 26.7\% \\
Git / DevOps / Kubernetes & 9 & 25.9\% & 33.3\% & 22.2\% \\
Travel / Planning & 7 & 23.8\% & 42.9\% & 14.3\% \\
Research / ML Datasets & 12 & 19.4\% & 25.0\% & 16.7\% \\
Other & 20 & 16.7\% & 25.0\% & 10.0\% \\
Monitoring / Operations & 5 & 13.3\% & 20.0\% & 0.0\% \\
Notion & 4 & 8.3\% & 25.0\% & 0.0\% \\
\bottomrule
\end{tabular}
\vspace{5pt}
\caption{Toolathlon performance by task category for the trained model. Categories are sorted by Pass@1. The strongest performance appears in Data Analysis/Finance and Web/Forms/Documents, while Monitoring/Operations and Notion remain more challenging.}
\label{tab:toolathlon_categories}
\end{table}

The increase in average turns per task from 27.9 (baseline) to 33.2 (trained) suggests that the trained model engages in more thorough exploration, making additional tool calls to verify or refine its outputs, reflecting a more deliberate problem-solving strategy consistent with the improved multi-step workflow execution observed in \corecraft{} evaluations (Section~\ref{sec:results}).

\paragraph{Implications for Transfer.} The Toolathlon results provide several insights about the nature of transfer from \corecraft{} training. First, gains are broadest in categories that share structural properties with \corecraft{} tasks (multi-step data retrieval, constraint satisfaction, structured output generation) rather than domain content. This is consistent with our hypothesis that \corecraft{} teaches generalizable workflow patterns rather than customer-support-specific knowledge. Second, the near-doubling of Pass$^3$ suggests that training improves not just peak capability but also reliability, an outcome particularly relevant for production deployment where consistency matters more than occasional success. Third, the weakness in Monitoring/Operations and Notion tasks highlights that certain capabilities may require targeted training environments, motivating future work on multi-domain curricula (Section~\ref{sec:discussion}).

\section{Discussion: Environment Design and Its Implications for Agent Training}
\label{sec:discussion}
The current production agent deployment landscape shows that reliability remains the main barrier to real-world adoption~\citep{pan2025measuring}. Our results suggest that the design properties of training environments play an important role in addressing this challenge.

Many existing agent benchmarks rely on synthetic data generation to achieve scale~\citep{qin2024toolllm}. While synthetic approaches enable large task counts, they may introduce distributional artifacts that limit transfer. The \corecraft{} environment takes a different approach, grounding tasks in realistic professional workflows through three properties that we believe contribute to the observed generalization. First, \corecraft{} features diverse task types spanning information retrieval, communication, reasoning, and multi-step workflows, requiring agents to develop a broad repertoire of skills rather than narrow heuristics. This diversity of difficulty levels within the training distribution relates to insights from curriculum learning research~\citep{narvekar2020curriculum}, which has shown that exposure to varied task difficulties can improve learning outcomes in RL domains. Second, expert-authored rubrics decompose success into verifiable sub-criteria, providing a richer reward signal. Third, tasks are designed to mirror authentic workplace challenges, encouraging models to learn professional patterns (e.g., structured communication, constraint-aware data retrieval) that apply across domains.


Our results suggest that environment quality, diversity, and realism play a meaningful role in producing transferable capabilities.

\subsection{Future Work}

Our results motivate several directions for future research:

\paragraph{Extended Training and Scaling.} We trained for a single epoch. Extended training will reveal saturation dynamics and confirm whether additional epochs yield diminishing returns or continued improvement.

\paragraph{Base Model Generalization.} We trained and evaluated GLM~4.6. Testing whether similar gains emerge across different model families and scales would further validate that environment quality, rather than model-specific interactions, drives improvement.

\paragraph{Ablation Studies.} Direct comparison against degraded environment variants (synthetic tasks, simplified rubrics, reduced entity complexity) would quantify the contribution of each design principle to transfer performance.

\paragraph{Multi-Domain Curricula.} Combining \corecraft{} with other high-fidelity enterprise environments from Surge could develop broader agent capabilities while testing for positive transfer between domains.

\section{Conclusion}
\label{sec:conclusion}

We presented evidence that high-fidelity RL environments enable generalizable agent training. Using \corecraft{}, a realistic customer support environment, we showed that single-epoch GRPO training produces 11.39 percentage point improvement on held-out tasks and transfers to out-of-distribution benchmarks (+4.5\% BFCL Parallel, +7.4\% $\tau^2$-Bench Retail, +6.8\% Toolathlon).

Our results highlight the importance of environment quality, diversity, and realism in agent training. Effective training environments should feature diverse and challenging tasks designed by domain experts, rubric-based evaluation enabling reliable reward computation, and realistic workflow patterns that mirror genuine professional settings. These design principles may help bridge the gap between benchmark performance and production deployment reliability.

The trained model learned transferable competencies in task decomposition, constraint handling, and response structuring rather than environment-specific heuristics. This suggests that careful environment design can shape what agents learn, steering them toward robust capabilities applicable across domains.





\section{Acknowledgements}
We thank the Surge AI workforce for their essential contributions to this research. Domain experts designed workplace tasks with complexity and realistic edge cases that reflect actual operational challenges. Annotators populated the \corecraft{} environment with coherent entities, relationships, and data that approximate real-world e-commerce systems. Expert evaluators assessed model trajectories and identified failure patterns, enabling the systematic analysis presented in this work. We thank Tim Bauman for leading \corecraft{} engineering and John Stacy for conducting thorough evaluations and analyses on external benchmarks.

\bibliographystyle{plainnat}

\end{document}